%% file: main.tex
\documentclass[runningheads]{llncs}

\usepackage{eccv}

\usepackage{eccvabbrv}

\usepackage{graphicx}
\usepackage{booktabs}

\usepackage[accsupp]{axessibility}  

\usepackage{hyperref}

\usepackage{orcidlink}

\usepackage{multirow}

\begin{document}

\title{DiGS-Avatar: Single-Image Animatable 3D Human Reconstruction via UV-Space Diffusion} 

\titlerunning{DiGS-Avatar}

\author{
    Jiakun Li\inst{1}\thanks{Equal contribution.} \and 
    Li Fang\inst{1}\thanks{Equal contribution. Corresponding author.} \and 
    Hao Zhu\inst{2} \and 
    Fei Hu\inst{1} \and 
    Long Ye\inst{1} \and 
    Yuan Zhang\inst{1} \and 
    Jinyao Yan\inst{1}
}

\authorrunning{J. Li et al.}

\institute{Key Laboratory of Media Audio and Video (Communication University of China), Ministry of Education, Beijing 100024, China\\
\email{lijiakun@mails.cuc.edu.cn, lifang8902@cuc.edu.cn, hufei@cuc.edu.cn}\\
\email{yelong@cuc.edu.cn, yuanzhang@cuc.edu.cn, jyan@cuc.edu.cn}
\and School of Intelligence Science and Technology, Nanjing University, Nanjing 210023, China\\
\email{zh@nju.edu.cn} }

\maketitle

\begin{abstract}
  Single-image 3D human reconstruction often suffers from over-smoothed textures and geometric inconsistencies. While diffusion models improve generative quality, their reliance on multi-view synthesis prior to 3D reconstruction is computationally expensive and prone to view inconsistency. We propose DiGS-Avatar, which reformulates this task as an efficient, diffusion-based UV-latent completion task, ensuring 3D consistency by design. To capture accurate spatial structure, we introduce a teacher-student framework where a multi-view teacher provides geometrically aligned pseudo-ground-truth latents to supervise a single-view diffusion student. Treating this inferred latent as a robust structural skeleton, our method injects high-level semantic features to accurately recover fine textural details without disrupting spatial integrity. The refined representation is then decoded into 3D Gaussian primitives. Extensive experiments demonstrate that DiGS-Avatar achieves state-of-the-art or highly competitive visual fidelity and zero-shot generalization, while reconstructing a fully animatable 3D avatar in just 0.71 seconds. Code is available at \href{https://github.com/KLMAV-CUC/DiGS-Avatar}{https://github.com/KLMAV-CUC/DiGS-Avatar}.
\end{abstract}

\section{Introduction}
\label{sec:intro}

\begin{figure}[tb]
  \centering
  \includegraphics[width=\linewidth]{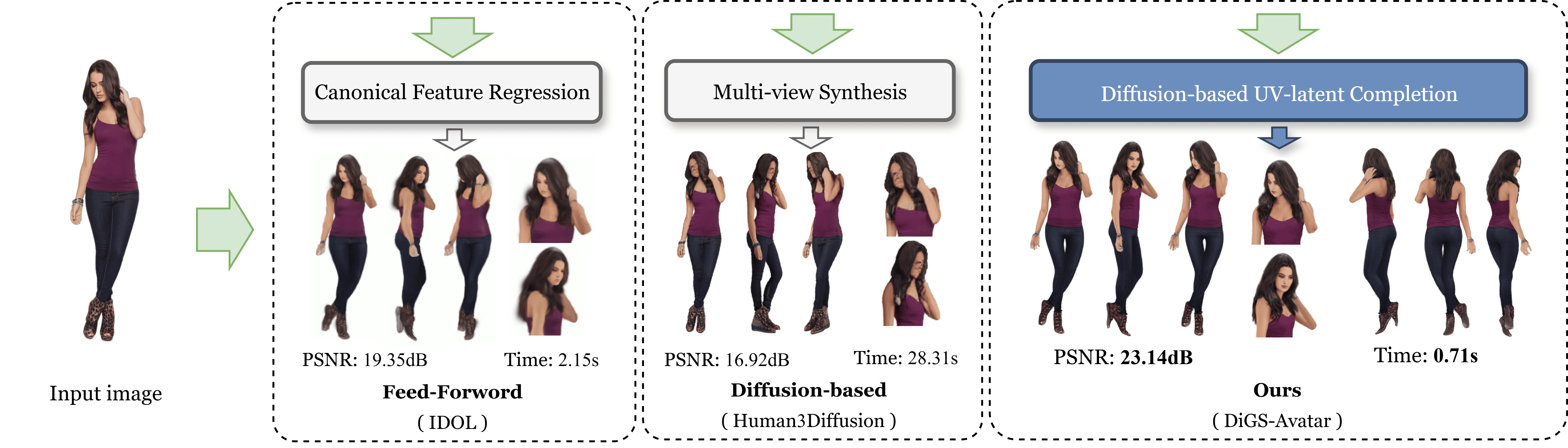}
  \caption{Comparison of paradigms for single-image 3D human reconstruction. (a) Feed-forward methods (e.g., IDOL \cite{zhuang2025idol}) directly regress 3D representations, enable fast inference but struggle with unseen regions and texture diversity. (b) Diffusion-based methods (e.g., Human3Diffusion \cite{xue2024human}) synthesize multi-view images before reconstruction, incurring high cost and view inconsistency. (c) Our \textbf{DiGS-Avatar} performs diffusion-based UV-latent completion in a canonical, surface-aligned space, achieving both photorealistic and geometrically consistent 3D avatars with efficient inference.
  }
  \label{fig:teaser}
\end{figure}

Reconstructing high-fidelity, animatable 3D humans from a single image is fundamental to applications such as virtual reality, gaming, telepresence, and 3D content creation. This task, however, is a highly ill-posed and challenging problem due to the complexity of clothing, the diversity of human poses, and severe occlusions.

To address the inherent ambiguity of single-view reconstruction, recent research has explored several paradigms. The first is a direct feed-forward approach, in which large models are trained to regress 3D representations directly from a single image. Some methods employ style-based encoders to generate triplanar representations \cite{chan2022efficient, hu2023sherf}, while others leverage large-scale synthetic datasets to train transformers that predict 3D Gaussian attributes \cite{kerbl20233d} in canonical space \cite{zhuang2025idol, qiu2025LHM}. However, these purely deterministic regressors lack strong generative priors, often producing over-smoothed geometry and struggling to hallucinate plausible details in unseen or occluded regions. A second line of research leverages the generative power of diffusion models \cite{ramesh2022hierarchical, rombach2022high}. Many methods synthesize multi-view representations---either rendered images or latent feature maps---before reconstructing 3D shapes \cite{pan2024humansplat, xue2024human, huang2025adahuman}, which is prone to subtle view inconsistencies. To bypass this, recent native 3D diffusion models \cite{xiang2025structured, chen2025synchuman} apply generative priors directly in 3D or triplane spaces. While these native 3D approaches achieve strict geometric consistency, they rely heavily on explicit 3D supervision and computationally expensive volumetric processing, resulting in slow training and inference speeds that hinder practical application.

In this work, we take a fundamentally different perspective on single-image 3D human reconstruction by reformulating it as an efficient, diffusion-based UV-latent completion task within a canonical, surface-aligned domain. Because every UV pixel maps to a unique surface point, our pure 2D formulation is 3D-consistent by design. Given an input image, features are first mapped to UV space using the SMPL-X model \cite{loper2015smpl, pavlakos2019expressive}, producing a partial UV latent that preserves spatial correspondence in visible regions. A 2D diffusion model is then utilized to complete this latent, inferring the missing texture and geometry. This approach effectively reduces the computationally heavy 3D generative process to a highly efficient 2D inpainting task, achieving a perfect balance between geometric consistency, generative flexibility, and inference speed.

Training a diffusion model for UV-latent completion requires structured supervision that is both geometrically consistent and semantically expressive. Since the complete, geometry-aligned UV-space latent corresponding to a single input image is unobservable, we introduce a novel teacher-student framework. A multi-view teacher is utilized to construct geometry-aligned, pseudo-ground-truth UV latents to supervise a single-view diffusion student. This enables the student to leverage a diffusion prior to infer a geometrically coherent UV latent directly from a single input image. Treating this inferred latent as a robust structural skeleton, we further propose the Geometry-Aligned Semantic Aggregation (GASA) module. GASA seamlessly injects high-level semantic features---extracted via a pretrained ViT-based appearance encoder---into the geometry-consistent skeleton. This ensures the accurate recovery of fine textural details without disrupting spatial integrity. The refined feature maps are finally decoded into a canonical 3D Gaussian Splatting (3DGS) \cite{kerbl20233d} representation for photorealistic rendering and animation.

Our main contributions are as follows:
\begin{itemize}
  \item We reformulate single-image 3D human reconstruction as an efficient 2D UV-latent completion task. Because every UV pixel maps to a unique surface point, this pure 2D approach is 3D-consistent by design, successfully bypassing both the computational bottlenecks of native 3D diffusion and the heavy synthesis costs of multi-view approaches.
  \item We implement this formulation via a teacher-student framework that enables robust 3D reconstruction using only 2D rendering supervision. This avoids inference-time multi-view image generation. Within this framework, a shared GASA module injects high-level semantic features to recover fine details, overcoming the texture loss inherent to VAE compression.
  \item Our model, DiGS-Avatar, generates photorealistic, fully rigged, and animatable 3D avatars from a single image in just 0.71 seconds. Extensive experiments show that it achieves state-of-the-art or highly competitive visual fidelity and zero-shot generalization with substantially higher inference efficiency.
\end{itemize}

\section{Related Work}
\label{sec:related}

\subsection{Single-Image 3D Human Reconstruction}
Single-image 3D human reconstruction has progressed from implicit-field representations to efficient, large-scale models capable of producing animatable results \cite{corona2023structured, hu2023sherf, huang2023one}. Early clothed human reconstruction methods such as PIFu \cite{saito2019pifu}, PIFuHD \cite{saito2020pifuhd}, ARCH \cite{huang2020arch, he2021arch++}, and PaMIR \cite{zheng2021pamir} adopt implicit surface representations but are computationally expensive and sensitive to SMPL estimation errors. Tex2Shape \cite{alldieck2019tex2shape} takes a different approach by directly regressing UV displacement and texture maps from a single image, providing pixel-aligned supervision in UV space but showing limited geometric detail and generalization ability. Hybrid approaches integrate explicit priors for improved robustness, including optimization-based methods \cite{xiu2022icon, xiu2023econ, zhang2023global, albahar2023single} and side-view-conditioned networks \cite{zhang2023global, zhang2024sifu}. Recent efforts focus on feed-forward reconstruction using expressive 3D representations. E$^3$Gen \cite{zhang20243gen} combines UV maps and Gaussian splatting \cite{kerbl20233d} to directly generate Gaussian attribute maps in UV space, though it is constrained in handling arbitrary image inputs. IDOL \cite{zhuang2025idol} learns UV-aligned 3DGS primitives from a single image, while LHM \cite{qiu2025LHM} employs a transformer-based architecture to decode multi-scale image tokens into 3DGS attributes in canonical space. These models offer fast inference and high-quality reconstruction but, as purely deterministic regressors, remain limited in hallucinating unseen geometry or texture owing to the inherent ambiguity of single-view input.

\subsection{Diffusion-Based 3D Human Reconstruction}
To address the ill-posed nature of single-view reconstruction, another active direction leverages the generative priors of diffusion models \cite{liu2023one, liu2024one, shi2023zero123++, long2024wonder3d, xu2024instantmesh, xudmv3d, zou2024triplane}. SIFU \cite{zhang2024sifu} refines UV-space textures via diffusion after an initial coarse reconstruction, improving appearance realism but remaining inefficient due to pixel-domain processing and an inability to correct underlying geometric errors. Human-LRM \cite{weng2024template} and GAS \cite{lu2025gas} adopt generalizable human NeRFs for canonical-space reconstruction followed by diffusion-based view or video refinement. HumanSplat \cite{pan2024humansplat}, PSHuman \cite{li2025pshuman}, and AniGS \cite{qiu2025anigs} extend this paradigm by generating multi-view latent, color, or normal maps for 3D or 4D Gaussian reconstruction. Human3Diffusion \cite{xue2024human} and AdaHuman \cite{huang2025adahuman} jointly integrate multi-view synthesis and 3DGS reconstruction. Dream, Lift, Animate \cite{buehler2026dream} similarly constructs animatable Gaussian avatars through multi-view generation and lifting, whereas DiGS-Avatar avoids inference-time multi-view synthesis by completing a single canonical UV latent. Although these multi-view approaches achieve impressive photorealism, generating representations in high-dimensional, view-dependent spaces increases computational cost and frequently introduces subtle geometric inconsistencies. 

To resolve these inconsistencies, recent approaches have turned to native 3D generative models. For instance, TRELLIS \cite{xiang2025structured} utilizes rectified flow models over a sparse 3D grid to generate structured 3D latents for versatile asset creation. Building on this 3D-native paradigm, SyncHuman \cite{chen2025synchuman} proposes a complex framework that synchronizes a 2D multiview generative model with a 3D structure generative model to enhance human reconstruction. While achieving strict geometric consistency, these methods rely heavily on computationally expensive 3D volumetric processing and multi-stage pipelines, resulting in slow training and inference speeds (e.g., SyncHuman requires 51 seconds per image). In contrast, our method reformulates the task as diffusion-based completion in a structured, canonical 2D UV latent space. By utilizing a multi-view teacher to distill geometrically aligned pseudo-ground-truth latents into a single-view student, we bypass both the view-inconsistencies of multi-view synthesis and the computational bottlenecks of native 3D generation, achieving highly efficient, 3D-consistent reconstruction in under a second.

\section{Method}
\label{sec:method}

\subsection{Overview}
\label{sec:overview}

The goal of DiGS-Avatar is to reconstruct a photorealistic and animatable 3D human from a single image. We address this highly ill-posed task by reformulating single-image 3D reconstruction as an efficient, diffusion-based UV-latent completion task within a canonical, surface-aligned domain. Because every UV pixel maps to a unique surface point, this pure 2D formulation is 3D-consistent by design. To achieve this, we introduce a novel teacher-student framework, where a multi-view teacher produces geometry-aligned, pseudo-ground-truth UV latents to supervise a single-view diffusion student.

\begin{figure}[tb]
  \centering
  \includegraphics[width=\linewidth]{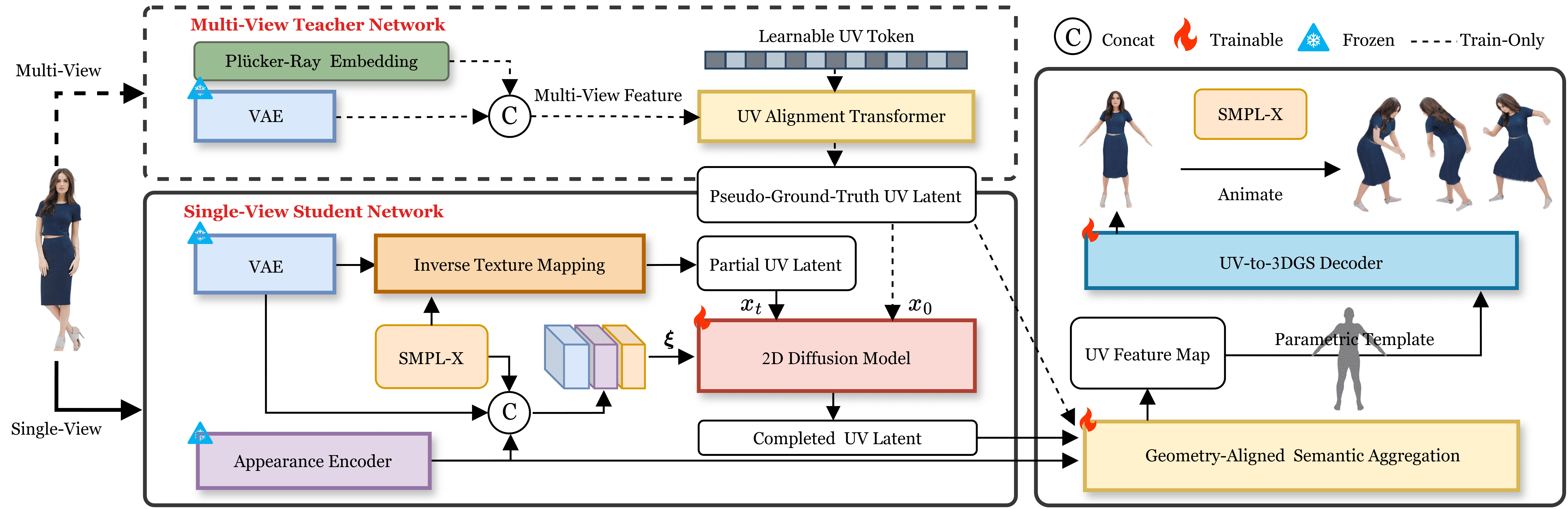}
  \caption{Overview of the proposed \textbf{DiGS-Avatar} framework. Given a single input image, the student network extracts VAE and appearance features. The VAE features are mapped to the canonical UV space via SMPL-X-based inverse texture mapping, forming a partial UV latent that is completed by a diffusion model under teacher supervision. The completed UV latent is enriched by GASA with appearance features, producing an aggregated UV feature map that is decoded into 3D Gaussian primitives for rendering and animation. During training, the multi-view teacher network provides geometrically consistent UV latents as supervision for the student.
  }
  \label{fig:pipeline}
\end{figure}

As shown in \cref{fig:pipeline}, the overall pipeline consists of three stages:

\textbf{Stage 1 -- Teacher Training:} The teacher network is trained using multi-view image sets. For each subject, it extracts and fuses features from multiple viewpoints into a single, complete, and view-consistent UV latent. This high-quality latent serves as the pseudo-ground-truth structural skeleton for the student model.

\textbf{Stage 2 -- Student Training:} The student network, built upon a 2D latent diffusion model, performs UV-latent completion from a single input image. It maps extracted features to the visible regions of the UV space via SMPL-X-based inverse texture mapping, and leverages a diffusion prior to infer the missing geometry and texture under the strict supervision of the teacher's latents.
    
\textbf{Stage 3 -- Decoding and Inference:} Both the teacher and student networks share a unified decoding pipeline. The completed UV latent is processed by our GASA module, which seamlessly injects high-level semantic appearance features into the latent structural skeleton to recover fine textural details. At inference time, only the lightweight single-view student path and the shared decoder are executed, outputting a canonical 3DGS avatar.

Following IDOL \cite{zhuang2025idol}, the reconstructed human is represented as a set of 3D Gaussian Splatting (3DGS) \cite{kerbl20233d} primitives parameterized directly in the canonical UV space of the SMPL-X \cite{pavlakos2019expressive} parametric model. Specifically, governed by shape ($\boldsymbol{\beta}$) and pose ($\boldsymbol{\theta}$) parameters, the SMPL-X template provides a continuous 2D UV parameterization where every coordinate uniquely maps to a physical point on the body surface. By predicting the 3DGS attributes---mean position $\mu \in \mathbb{R}^{3}$, scale $s \in \mathbb{R}^{3}$, rotation $r \in \mathbb{R}^{4}$, opacity $\rho \in [0,1]$, and appearance $c \in \mathbb{R}^{C}$---strictly within this geometry-aligned UV space, our formulation provides a canonical, surface-aligned representation that inherently supports immediate animation via standard Linear Blend Skinning. Unlike previous works that attempt to refine UV maps in the high-dimensional pixel domain \cite{alldieck2019tex2shape, zhang2024sifu} or rely on computationally heavy 3D volumetric diffusion, our method performs generative completion entirely within a structured 2D latent space. This successfully bridges the gap between powerful 2D generative priors and explicit 3D representations, enabling the generation of a fully rigged avatar in just 0.71 seconds.

\subsection{Network Architecture}
\label{sec:arch}
Our architecture shares the image encoders, the GASA module, and the UV-to-3DGS decoder across both the teacher and student networks to ensure a consistent representation space. The key distinction lies in how the complete UV latent is obtained: the teacher synthesizes it from multi-view images, while the student learns to complete it from a single image under teacher supervision.

\subsubsection{Feature Extraction.}
We adopt a VAE \cite{pu2016variational} encoder to provide a compact, geometry-consistent structural skeleton suitable for diffusion. However, its highly compressed spatial resolution (e.g., $32 \times 32$ with $32$ channels) inherently loses high-frequency details. To overcome this limitation, we introduce a pretrained ViT-based appearance encoder (implemented with DINOv3 \cite{simeoni2025dinov3}) to extract rich, high-dimensional semantic features (e.g., $768$ channels). The VAE and appearance encoders thus play complementary roles: the VAE supplies a geometrically coherent foundation, while the appearance encoder provides the semantic material needed to hallucinate fine details.

\subsubsection{Multi-View Teacher Network.}
Given $N$ multi-view images $\{ \boldsymbol{I}_i \}_{i=1}^N$ with known camera parameters, the teacher fuses the extracted VAE latents, $\{ \boldsymbol{F}_i^\text{VAE} \}_{i=1}^N$, using a UV-Alignment Transformer. A set of learnable UV tokens is concatenated with the flattened per-view VAE features. Crucially, Plücker ray embeddings \cite{plucker1865xvii} are appended to these features to explicitly encode both the camera origin and ray direction. Processed through $D$ transformer blocks with multi-head self-attention, this spatial conditioning forces the network to resolve multi-view inconsistencies, outputting a strictly geometry-aligned pseudo-ground-truth UV latent, $\boldsymbol{Z}^\text{VAE}$, for supervising the student network.

\subsubsection{Single-View Student Network.}
Once trained, the teacher network provides the complete UV latents that serve as supervision targets. The student network learns to reconstruct this same complete latent given only a single reference image $\boldsymbol{I}_\text{ref}$, which is passed through both the VAE and the appearance encoder to obtain features $\boldsymbol{F}_\text{ref}^\text{VAE}$ and $\boldsymbol{F}_\text{ref}^\text{app}$, respectively. Unlike the teacher, the student maps the single-view VAE features $\boldsymbol{F}_\text{ref}^\text{VAE}$ to the UV space via SMPL-X-based inverse texture mapping, resulting in a partial UV latent that covers only the visible regions of the human body.

To complete the missing regions, we employ a conditional 2D diffusion model $D_\theta$ \cite{xie2025sana}. Given the teacher-provided ground-truth latent $\boldsymbol{x}_0 = \boldsymbol{Z}^\text{VAE}$, we randomly sample a timestep $t$ and generate its noisy version:
\begin{equation}
    \boldsymbol{x}_t = \sqrt{\bar{\alpha}_t} \boldsymbol{x}_0 + \sqrt{1 - \bar{\alpha}_t}\,\epsilon, 
    \quad \epsilon \sim \mathcal{N}(0, \mathbf{I}),
\label{eq:diff-step}
\end{equation}
where $\bar{\alpha}_t$ follows a cosine noise schedule. The diffusion model predicts the noise $\hat{\epsilon}_\theta$ to recover the clean latent $\boldsymbol{x}_0$ from $\boldsymbol{x}_t$, conditioned on the fused geometric and appearance context:
\begin{equation}
    \boldsymbol{\xi} = \boldsymbol{F}_\text{ref}^\text{VAE} \oplus \boldsymbol{F}_\text{ref}^\text{app} \oplus \boldsymbol{N}_\text{ref}^\text{SMPL-X},
\label{eq:condition}
\end{equation}
where $\oplus$ denotes channel-wise concatenation and $\boldsymbol{N}_\text{ref}^\text{SMPL-X}$ represents the surface normal map rendered from the estimated SMPL-X mesh at the reference viewpoint. This explicit 2D spatial conditioning enables the diffusion model to infer the missing UV content by relying on the geometric structure from the VAE, the detailed appearance cues from the appearance encoder, and the dense topological priors provided by the normal map. At inference, we employ a deterministic sampling scheme to efficiently predict the completed UV latent $\boldsymbol{Z}^\text{VAE} = \hat{\boldsymbol{x}}_0$.

\subsubsection{GASA Module.}
While the geometry-aligned UV latent $\boldsymbol{Z}^\text{VAE}$ provides a robust 3D-consistent skeleton, its $32 \times 32$ spatial resolution is insufficient for high-fidelity rendering. To bridge this gap, the GASA module employs a progressive cross-attention mechanism to inject the multi-level appearance features $\boldsymbol{F}_\text{ref}^\text{app}$ into the structural skeleton.

Specifically, we divide appearance features into three groups: deep features $\boldsymbol{F}^{\text{app}}_{\text{deep}}$, middle features $\boldsymbol{F}^{\text{app}}_{\text{mid}}$, and shallow features $\boldsymbol{F}^{\text{app}}_{\text{shallow}}$. GASA then performs coarse-to-fine semantic aggregation in three spatial stages. First, it aggregates the deep-level features into the UV latent at the base $32 \times 32$ resolution via cross-attention, where $\boldsymbol{Z}^\text{VAE}$ serves as the query and $\boldsymbol{F}_\text{deep}^\text{app}$ provides the keys/values:
\begin{equation}
    \boldsymbol{Z}^{(1)} = \text{CrossAttn}_{32}(\boldsymbol{Z}^\text{VAE}, \boldsymbol{F}_\text{deep}^\text{app}).
\label{eq:gasa_stage1}
\end{equation}
Second, this enriched latent is spatially upsampled to a $64 \times 64$ resolution and subjected to a subsequent round of cross-attention to inject mid-level features:
\begin{equation}
    \boldsymbol{Z}^{(2)} = \text{CrossAttn}_{64}(\text{Upsample}(\boldsymbol{Z}^{(1)}), \boldsymbol{F}_\text{mid}^\text{app}).
\label{eq:gasa_stage2}
\end{equation}
Finally, the latent is upsampled to $128 \times 128$ and queries shallow-level features, which preserve richer local texture cues:
\begin{equation}
    \boldsymbol{F}_\text{UV} = \text{CrossAttn}_{128}(\text{Upsample}(\boldsymbol{Z}^{(2)}), \boldsymbol{F}_\text{shallow}^\text{app}).
\label{eq:gasa_stage3}
\end{equation}
This progressive injection ensures that the final high-resolution UV feature map $\boldsymbol{F}_\text{UV}$ successfully recovers fine textural details while strictly adhering to the geometry-aligned constraints of the VAE latent. This shared module ensures that both teacher supervision and student inference operate within a unified, geometry-aware appearance space.

\subsubsection{UV-to-3DGS Decoder.}
The decoder transforms the UV feature map $\boldsymbol{F}_\text{UV}$ into multi-channel Gaussian attribute maps:
\begin{equation}
    \{\boldsymbol{M}_{\text{pos}}, \boldsymbol{M}_{\text{rot}}, \boldsymbol{M}_{\text{sca}}, \boldsymbol{M}_{\text{col}}, \boldsymbol{M}_{\text{opa}}\} = \text{Dec}_{3\text{DGS}}(\boldsymbol{F}_\text{UV}),
\label{eq:dec3dgs}
\end{equation}
which are sampled to obtain the Gaussian primitives $G_k = \{ \boldsymbol{\mu}_k, \rho_k, \boldsymbol{r}_k, \boldsymbol{s}_k, \boldsymbol{c}_k \}$. To ensure topological consistency, each Gaussian’s mean, scale, and rotation are predicted as explicit offsets relative to its corresponding SMPL-X vertex:
\begin{equation}
    \boldsymbol{\mu}_k = \hat{\boldsymbol{\mu}}_k + \delta \boldsymbol{\mu}_k,\quad \boldsymbol{s}_k = \hat{\boldsymbol{s}}_k \odot \delta \boldsymbol{s}_k,\quad \boldsymbol{r}_k = \hat{\boldsymbol{r}}_k \cdot \delta \boldsymbol{r}_k,
\label{eq:gauss_relative}
\end{equation}
where $\hat{\boldsymbol{\mu}}_k$, $\hat{\boldsymbol{s}}_k$, and $\hat{\boldsymbol{r}}_k$ denote the canonical SMPL-X parameters, and ``$\odot$'' denotes element-wise multiplication. Finally, the canonical Gaussians are animated via Linear Blend Skinning (LBS):
\begin{equation}
    \boldsymbol{\mu}'_k = \sum_{i=1}^{n_b} w_i \mathbf{B}_i \boldsymbol{\mu}_k,
\label{eq:lbs}
\end{equation}
with $w_i$ and $\mathbf{B}_i$ representing the skinning weights and per-joint transformations. This structured decoding process produces an explicit, pose-aware 3D Gaussian avatar that is both photorealistic and instantly animatable.

\subsection{Training Objectives}
\label{sec:loss}

\subsubsection{Teacher Loss.}
The multi-view teacher network is trained end-to-end with the shared GASA module and the UV-to-3DGS decoder to establish a high-quality, view-consistent latent space. Following IDOL \cite{zhuang2025idol}, we supervise the teacher using a photometric rendering objective that combines a Mean Squared Error (MSE) loss, a VGG perceptual loss, and a Gaussian position regularization term:
\begin{equation}
    \mathcal{L}_\text{teacher} = \mathcal{L}_\text{photo} = \mathcal{L}_\text{MSE} + \lambda_\text{vgg} \mathcal{L}_\text{VGG} + \lambda_\text{off} \mathcal{L}_\text{Offset}.
\label{eq:rec_loss}
\end{equation}
Here, $\mathcal{L}_\text{MSE}$ and $\mathcal{L}_\text{VGG}$ are computed between the rendered novel views and the ground-truth multi-view images. The regularization term $\mathcal{L}_\text{Offset} = \| \delta \boldsymbol{\mu}_k \|_2^2$ penalizes large deviations of the predicted Gaussian centers from their canonical SMPL-X vertex positions, ensuring the topological stability.

\subsubsection{Student Loss.}
The student diffusion model $D_\theta$ learns to complete the missing regions in the UV latent, using the clean UV latent produced by the frozen teacher ($\boldsymbol{x}_0 = \boldsymbol{Z}^\text{VAE}$) as the pseudo-ground-truth target. The primary denoising objective follows the standard latent diffusion formulation:
\begin{equation}
    \mathcal{L}_\text{diff} = \mathbb{E}_{\boldsymbol{x}_0, \epsilon, t} \left[ \| \epsilon - \epsilon_\theta(\boldsymbol{x}_t, t, \boldsymbol{\xi}) \|_2^2 \right],
\label{eq:diff_loss}
\end{equation}
where $\boldsymbol{x}_t$ is the noisy latent at timestep $t$, and $\epsilon_\theta$ denotes the predicted noise by the diffusion model, conditioned on $\boldsymbol{\xi}$ in \cref{eq:condition}. To ensure appearance and geometric fidelity, we also compute a photometric loss $\mathcal{L}_\text{photo}$ between the student’s rendered views and the ground truth. The total student objective is therefore:
\begin{equation}
    \mathcal{L}_\text{student} = \mathcal{L}_\text{diff} + \lambda_\text{photo} \mathcal{L}_\text{photo}.
\label{eq:student_loss}
\end{equation}
This joint formulation ensures that the student accurately learns the 2D diffusion prior while remaining strictly anchored to the 3D-consistent rendering space defined by the teacher.

\section{Experiments}
\label{sec:exp}

\subsection{Implementation Details}
\label{sec:impl}

\subsubsection{Datasets and Evaluation.}
We utilize HuGe100K \cite{zhuang2025idol}, THuman 2.1 \cite{zheng2019deephuman} (72 spherical views/subject), and 2K2K \cite{han2023high} (24 yaw-axis views/subject), rendered at $896 \times 640$ resolution. The last 50 subjects per dataset form the test set. The most frontal view serves as the reference input, with unseen views used as ground-truth targets for photometric evaluation. All methods receive identical camera and ground-truth SMPL-X parameters to ensure fair comparison. For zero-shot generalization, we evaluate on SIZER \cite{tiwari2020sizer} and in-the-wild images (DeepFashion \cite{liu2016deepfashion}, Internet), extracting SMPL-X parameters for the latter via Multi-HMR \cite{baradel2024multi}.

\subsubsection{Training Procedure.}
Implemented in PyTorch, DiGS-Avatar trains on a single NVIDIA A100 GPU (batch size 1, Adam optimizer \cite{kingma2014adam}, learning rate $5 \times 10^{-4}$). The appearance encoder utilizes the pre-trained DINOv3 model \cite{simeoni2025dinov3}, whose weights remain strictly frozen throughout the entire process. Training follows a staged pipeline. First, the teacher network trains for 60K iterations on HuGe100K and THuman 2.1 (\cref{eq:rec_loss}), taking the frontal view plus 4 random views as input. To prevent identity mapping, it is supervised by 4 distinct novel views. Next, we initialize the student with the teacher's frozen VAE, GASA, and UV-to-3DGS decoder, and optimize only the diffusion model for 20K iterations with a 2K-step warm-up. Finally, to improve multi-view consistency and reduce geometric noise from HuGe100K, we fine-tune the teacher for 45K iterations and continue student diffusion training with rendering supervision for 15K iterations on the high-fidelity THuman 2.1 and 2K2K datasets. We use a Soft Blending Euler sampler with 20 steps and CFG=0.1. The loss weights are $\lambda_\text{vgg}=1.0$, $\lambda_\text{off}=0.5$, and $\lambda_\text{photo}=0.5$. The reported 0.71s inference time covers the full pipeline from input image to final 3DGS output.

\subsection{Comparison with Existing Methods}
\label{sec:comp}

\begin{table}[tb]
  \caption{Quantitative comparison on in-distribution datasets. “Ours” refers to the diffusion-based student network used during inference. The best result is highlighted in \textbf{bold} and the second best is \underline{underlined}. * denotes animatable representations.}
  \label{tab:comparison_main}
  \centering
  \resizebox{\linewidth}{!}{
  \begin{tabular}{lccccccccc}
    \toprule
    \multirow{2}{*}{Methods} & \multicolumn{3}{c}{HuGe100K \cite{zhuang2025idol}} & \multicolumn{3}{c}{THuman 2.1 \cite{zheng2019deephuman}} & \multicolumn{3}{c}{2K2K \cite{han2023high}} \\
    \cmidrule(r){2-4} \cmidrule(r){5-7} \cmidrule(r){8-10} 
    & PSNR $\uparrow$ & SSIM $\uparrow$ & LPIPS $\downarrow$ & PSNR $\uparrow$ & SSIM $\uparrow$ & LPIPS $\downarrow$ & PSNR $\uparrow$ & SSIM $\uparrow$ & LPIPS $\downarrow$ \\
    \midrule
    SIFU \cite{zhang2024sifu} & 18.44 & 0.921 & 0.095 & 16.82 & 0.926 & 0.095 & 18.54 & \textbf{0.931} & 0.091 \\
    Human3Diffusion \cite{xue2024human} & 18.09 & 0.893 & 0.101 & 15.67 & 0.902 & 0.105 & 19.29 & 0.908 & 0.079 \\
    IDOL* \cite{zhuang2025idol} & \underline{22.75} & \underline{0.926} & 0.084 & 18.82 & \underline{0.937} & 0.077 & 18.17 & 0.916 & 0.092 \\
    LHM* \cite{qiu2025LHM} & 21.76 & 0.919 & \underline{0.065} & \underline{19.62} & 0.931 & \underline{0.072} & \underline{20.67} & 0.918 & 0.073 \\
    TRELLIS \cite{xiang2025structured} & 18.24 & 0.895 & 0.093 & 16.52 & 0.910 & 0.094 & 18.72 & 0.907 & 0.084 \\
    SyncHuman \cite{chen2025synchuman} & 17.52 & 0.893 & 0.092 & 17.04 & 0.911 & 0.089 & 20.27 & 0.917 & \underline{0.069} \\
    \midrule
    Ours* & \textbf{25.32} & \textbf{0.940} & \textbf{0.039} & \textbf{21.45} & \textbf{0.938} & \textbf{0.060} & \textbf{22.22} & \underline{0.919} & \textbf{0.064} \\
  \bottomrule
  \end{tabular}
  }
\end{table}

\subsubsection{Baselines.}
We compare DiGS-Avatar against representative state-of-the-art methods across four paradigms: deterministic feed-forward regressors (IDOL \cite{zhuang2025idol}, LHM \cite{qiu2025LHM}), pixel-domain texture diffusion (SIFU \cite{zhang2024sifu}), multi-view diffusion (Human3Diffusion \cite{xue2024human}), and native 3D generative models (TRELLIS \cite{xiang2025structured}, SyncHuman \cite{chen2025synchuman}). As discussed in \cref{sec:related}, while these baselines excel in specific areas, they often struggle to balance high-frequency generative detail, strict geometric consistency, and inference efficiency. To ensure the fairest possible comparison, and owing to the prohibitive computational resources required to retrain massive volumetric baselines from scratch on our combined data splits (e.g., IDOL alone requires 32 H100 GPUs), we strictly evaluate all methods utilizing their officially released pre-trained weights and recommended inference pipelines.

\subsubsection{Comparison with State-of-the-Art.}

\begin{figure}[tb]
  \centering
  \includegraphics[width=\linewidth]{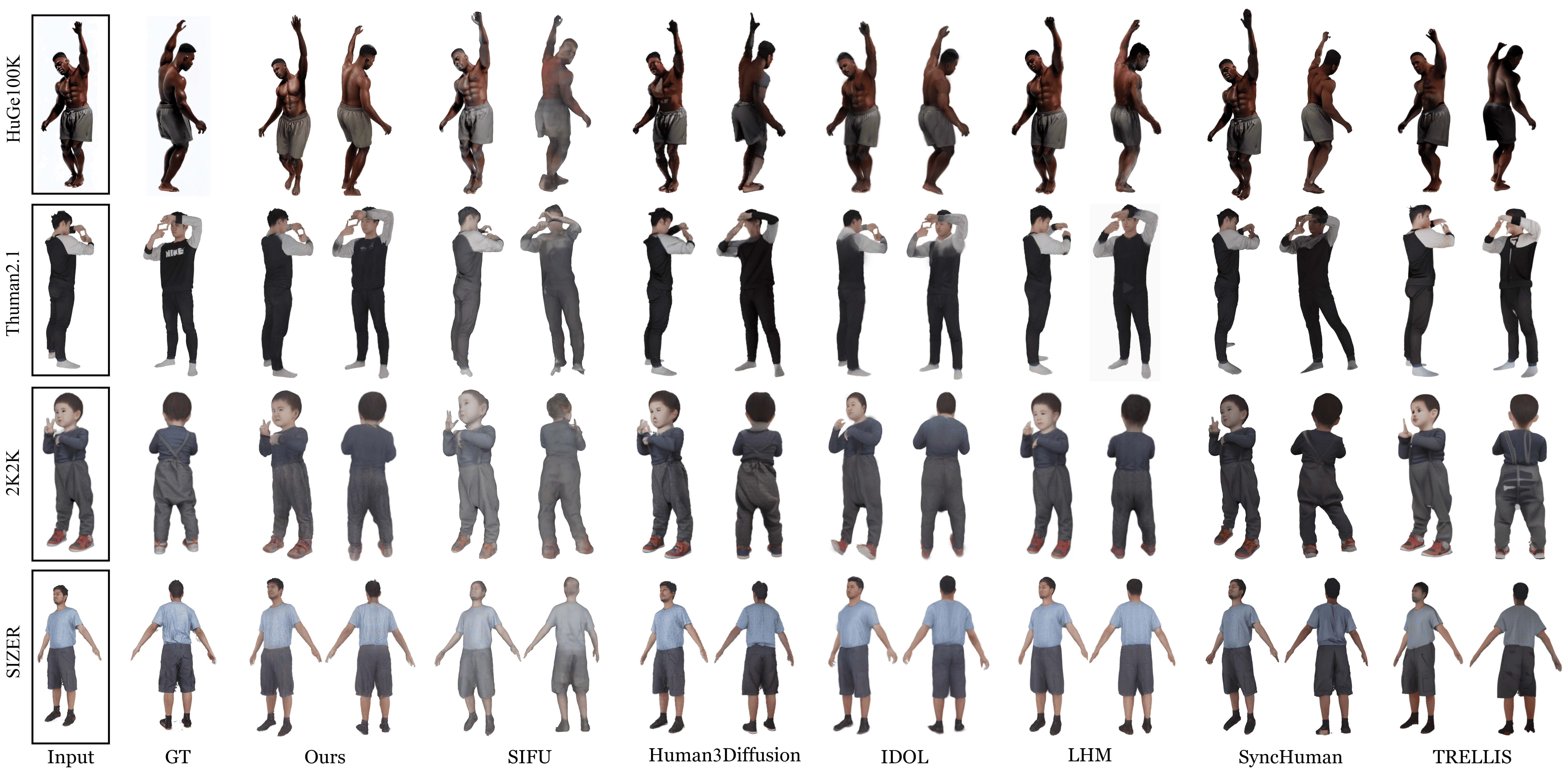}
  \caption{Qualitative comparison on datasets (top-down): HuGe100K, THuman 2.1, 2K2K, and SIZER. Deterministic baselines over-smooth unseen regions, while multi-view diffusion methods exhibit geometric inconsistencies. In contrast, DiGS-Avatar produces sharp, 3D-consistent details. Notably, on the out-of-distribution SIZER dataset (bottom), our method uniquely preserves the complex geometry of loose clothing.}
  \label{fig:dataset_results}
\end{figure}

We evaluate DiGS-Avatar on the test splits of HuGe100K, THuman 2.1, and 2K2K using PSNR, SSIM \cite{wang2004image}, and LPIPS \cite{zhang2018unreasonable} metrics. As shown in \cref{tab:comparison_main}, DiGS-Avatar consistently outperforms the baselines. On HuGe100K and THuman 2.1, our method significantly surpasses feed-forward regressors like IDOL \cite{zhuang2025idol} and LHM \cite{qiu2025LHM} across all metrics, demonstrating the superiority of our generative UV-completion prior for recovering high-fidelity textures. On the 2K2K dataset, DiGS-Avatar achieves the highest PSNR and remains highly competitive in SSIM and LPIPS.

Qualitatively (\cref{fig:dataset_results}, top three rows), while deterministic baselines suffer from over-smoothed side or back views and multi-view diffusion models exhibit geometric inconsistencies, DiGS-Avatar generates sharp, high-frequency textures (e.g., fabric folds, facial features, and clothing patterns) while maintaining strict 3D consistency. This visual fidelity clearly demonstrates the advantage of our surface-aligned UV-latent completion formulation combined with the semantic enrichment of the GASA module.

Furthermore, our pure 2D formulation yields massive efficiency gains (\cref{tab:sizer_time}). With a 0.71s inference time, DiGS-Avatar is over 3$\times$ faster than the most efficient regressor (IDOL) and about 70$\times$ faster than native 3D generative models (SyncHuman). Coupled with a $\sim$60 GPU-hour training cost, it achieves a strong balance of visual fidelity and efficiency.

\subsubsection{Zero-Shot Generalization and In-the-Wild Evaluation.}
\label{sec:zeroshot}

To demonstrate the robustness of our learned generative priors against overfitting, we perform zero-shot evaluation on SIZER \cite{tiwari2020sizer}. SIZER is a challenging zero-shot benchmark with substantial clothing-size and loose-garment variation, directly stressing SMPL-X-bound topology. Quantitatively (\cref{tab:sizer_time}), DiGS-Avatar achieves the highest PSNR and SSIM scores, remaining highly competitive perceptually (LPIPS). Qualitatively (\cref{fig:dataset_results}, bottom row), while deterministic baselines struggle with baggy garments, our model successfully hallucinates plausible structural details and preserves the geometric integrity of the oversized clothing. This supports strong cross-dataset generalization rather than dataset memorization.

\begin{table}[tb]
  \caption{Zero-shot generalization on SIZER and computational efficiency. DiGS-Avatar achieves competitive zero-shot quality while requiring substantially lower inference and training cost.}
  \label{tab:sizer_time}
  \centering
  \resizebox{\linewidth}{!}{
  \begin{tabular}{lccccc}
    \toprule
    \multirow{2}{*}{Methods} & \multicolumn{3}{c}{SIZER \cite{tiwari2020sizer}} & \multicolumn{2}{c}{Computational Cost} \\
    \cmidrule(r){2-4} \cmidrule(r){5-6}
    & PSNR $\uparrow$ & SSIM $\uparrow$ & LPIPS $\downarrow$ & Train Time (in GPU hours) & Test Time (s) $\downarrow$ \\
    \midrule
    SIFU \cite{zhang2024sifu} & 19.30 & 0.905 & 0.101 & - & 30 \\
    Human3Diffusion \cite{xue2024human} & 19.14 & 0.904 & 0.088 & $\sim$960 & 28 \\
    IDOL* \cite{zhuang2025idol} & 21.90 & 0.922 & 0.075 & $\sim$768 & \underline{2.25} \\
    LHM* \cite{qiu2025LHM} & \underline{23.45} & \underline{0.929} & \textbf{0.064} & $\sim$2,496 & 5.13 \\
    TRELLIS \cite{xiang2025structured} & 19.94 & 0.916 & 0.078 & $\sim$10k & 7.43 \\
    SyncHuman \cite{chen2025synchuman} & 19.56 & 0.916 & 0.074 & - & 51 \\
    \midrule
    Ours* & \textbf{23.54} & \textbf{0.932} & \underline{0.065} & $\sim$60 & \textbf{0.71} \\
  \bottomrule
  \end{tabular}
  }
\end{table}

\begin{figure}[tb]
  \centering
  \includegraphics[width=\linewidth]{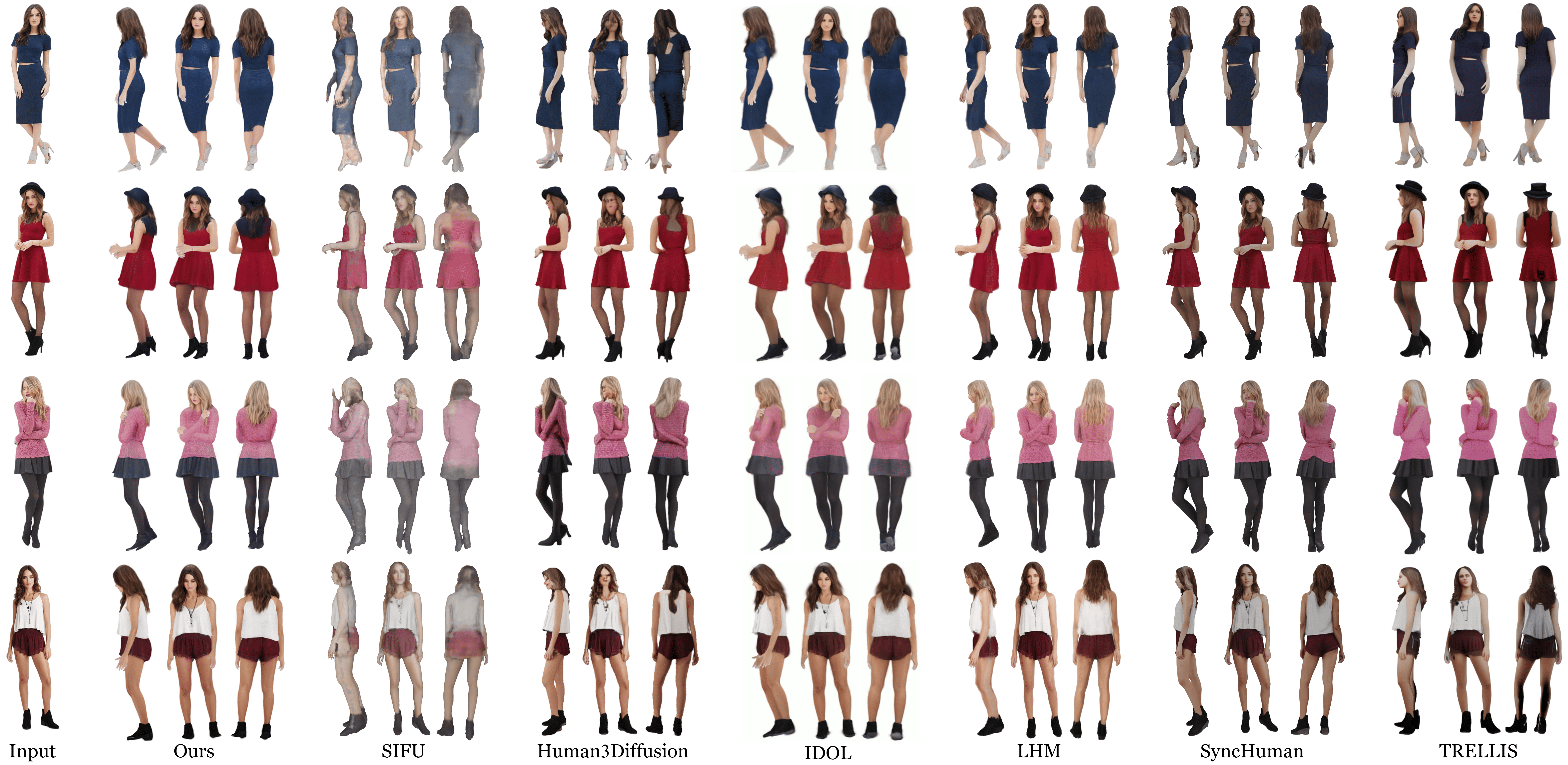}
  \caption{Zero-shot reconstruction on in-the-wild images from DeepFashion \cite{liu2016deepfashion} and the Internet. Using Multi-HMR \cite{baradel2024multi} for SMPL-X estimation, our model successfully handles complex real-world illumination, diverse poses, and layered casual clothing.}
  \label{fig:inthewild_results}
\end{figure}

For practical applicability, we evaluate unconstrained in-the-wild images from DeepFashion \cite{liu2016deepfashion} and the Internet, leveraging Multi-HMR \cite{baradel2024multi} to estimate SMPL-X parameters. As illustrated in \cref{fig:inthewild_results}, DiGS-Avatar reconstructs detailed, animatable avatars with diverse casual clothing and complex real-world illumination, demonstrating robust real-world generalization.

\subsection{Animation and Novel Pose Synthesis}
\label{sec:animation}

\begin{figure}[tb]
  \centering
  \includegraphics[width=\linewidth]{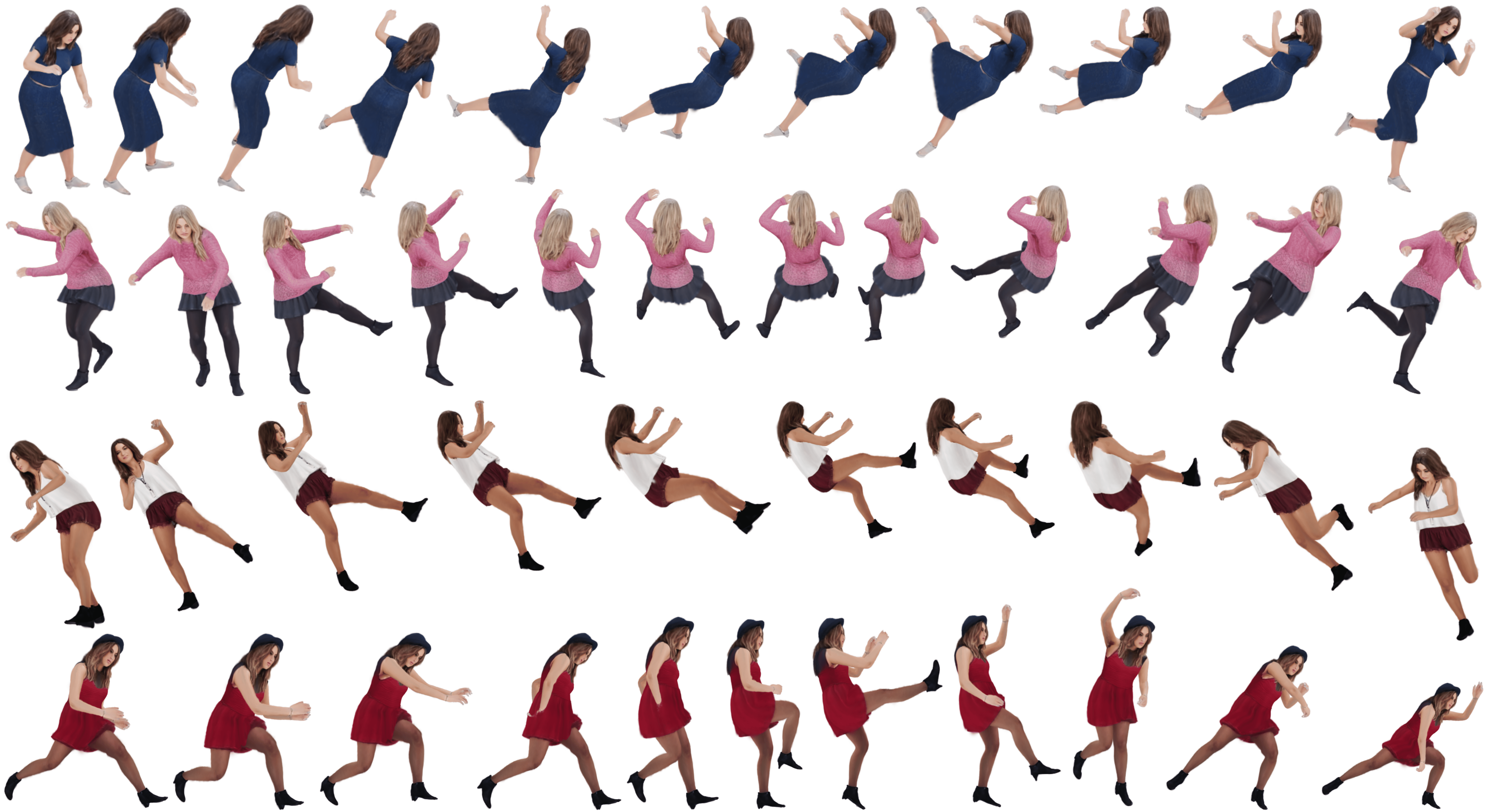}
  \caption{Animation results. Reconstructed avatars are directly driven by LBS and preserve coherent texture attachment under novel poses.}
  \label{fig:animation}
\end{figure}

Our canonical UV-latent formulation yields instantly animatable 3D avatars. Unlike implicit NeRFs or triplanar methods that require complex learned deformation fields, DiGS-Avatar explicitly binds the 3D Gaussian primitives to the SMPL-X template. Thus, standard LBS directly drives the avatars in novel poses without post-processing or test-time optimization. As shown in \cref{fig:animation}, driven by dynamic motion sequences, our reconstructions maintain strict structural integrity and coherent textures even under extreme articulations (e.g., complex dance movements). Because GASA deeply injects semantic features into the geometry-aligned UV space, high-frequency details---like intricate clothing patterns and facial features---remain strictly coherent. Consequently, DiGS-Avatar avoids the "texture sliding" and geometric tearing common in purely implicit models. Please refer to our supplementary video for temporal evaluations.

\subsection{Ablation Study and Analysis}
\label{sec:ablation}

We conduct ablation studies to assess the contribution of our key architectural components. The quantitative results are summarized in \cref{tab:ablations}, and qualitative comparisons are visualized in \cref{fig:ablation} (left). 

\begin{table}[t]
\caption{Quantitative ablation of key components on HuGe100K. ``Ours'' refers to the diffusion-based student network used during inference.}
\label{tab:ablations}
\centering
\begin{tabular}{lccc@{\hspace{0.8em}}lccc}
\toprule
Variant & PSNR $\uparrow$ & SSIM $\uparrow$ & LPIPS $\downarrow$ & Variant & PSNR $\uparrow$ & SSIM $\uparrow$ & LPIPS $\downarrow$ \\
\cmidrule(r){1-4} \cmidrule(r){5-8}
Teacher & 26.18 & 0.942 & 0.037 &
w/o Teacher & 19.03 & 0.804 & 0.168 \\
w/o GASA & 22.52 & 0.915 & 0.084 &
w/ LDM-UNet & 17.39 & 0.790 & 0.204 \\
w/o condition & 23.26 & 0.936 & 0.061 &
w/o $\mathcal{L}_\text{diff}$ & 15.98 & 0.785 & 0.226 \\
\midrule
Ours (full) & 25.32 & 0.940 & 0.039 & & & & \\
\bottomrule
\end{tabular}
\end{table}

\subsubsection{Effect of the Multi-View Teacher.}
The multi-view teacher provides the strongest supervision signal in our framework. As shown in \cref{tab:ablations}, the teacher achieves the best reconstruction quality (26.18 PSNR and 0.037 LPIPS), benefiting from multi-view observations during its forward pass. Our single-view student closely approaches this upper bound, achieving 25.32 PSNR and 0.039 LPIPS from only one input image. In contrast, removing teacher supervision causes a severe degradation to 19.03 PSNR and 0.168 LPIPS. This confirms that the teacher is not merely an auxiliary training component; it defines a geometry-aligned UV latent space that enables stable single-view diffusion completion.

\subsubsection{Effect of the Diffusion Prior and Denoising Loss.}
We further validate the necessity of diffusion-based UV-latent completion. Replacing our diffusion backbone with a standard LDM-UNet leads to a substantial drop to 17.39 PSNR and 0.204 LPIPS. This is because our compact $32 \times 32 \times 32$ UV latent has a very high compression ratio and requires strong global modeling and semantic priors. Removing the denoising loss $\mathcal{L}_\text{diff}$ degrades performance even further, yielding only 15.98 PSNR and 0.226 LPIPS. These results show that both the adapted diffusion prior and the denoising objective are essential for plausible and geometry-consistent UV-latent completion.

\subsubsection{Effect of the GASA Module.}
Removing GASA reduces PSNR from 25.32 to 22.52 and increases LPIPS from 0.039 to 0.084. Qualitatively, as shown in \cref{fig:ablation} (left), relying only on the compressed VAE latent produces blurry and over-smoothed reconstructions with weak high-frequency appearance. This demonstrates that GASA is critical for injecting rich DINOv3 appearance features into the geometry-aligned UV latent and recovering fine texture details.

\subsubsection{Effect of Conditioning.}
Removing the conditioning branch also clearly degrades reconstruction quality, reducing PSNR from 25.32 to 23.26 and increasing LPIPS from 0.039 to 0.061. Without explicit conditioning, the diffusion model has weaker guidance for aligning hallucinated UV content with the observed image evidence and SMPL-X topology. This confirms that conditioning is important for precise UV-latent completion, especially when the higher Gaussian budget allows finer details to be represented.

\subsubsection{Robustness to Extreme Poses and Topologies.}
\begin{figure}[t]
  \centering
  \includegraphics[width=\linewidth]{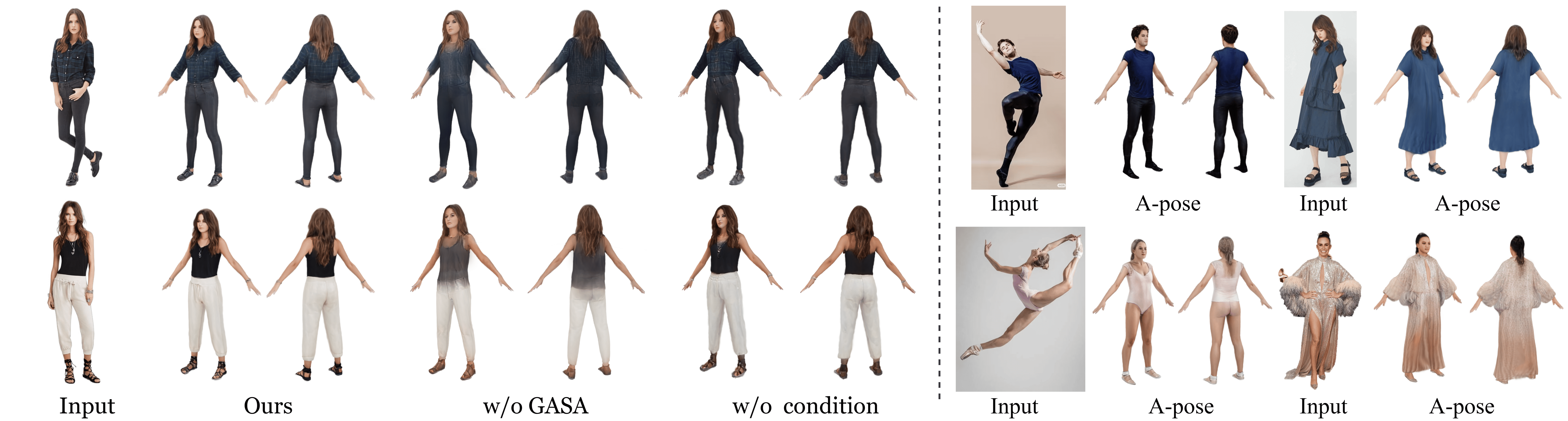}
  \caption{Ablation and Robustness Analysis. Left: Qualitative ablation of key components. Right: Robustness to challenging in-the-wild inputs.}
  \label{fig:ablation}
\end{figure}

Beyond architectural ablations, we further analyze robustness to challenging inputs. As shown in \cref{fig:ablation} (right), DiGS-Avatar handles extreme articulations and loose garments, such as leaping dancers, flowing dresses, and puffy skirts, while maintaining coherent canonical UV outputs. Nevertheless, because our representation remains anchored to the SMPL-X topology, extremely loose or non-rigid clothing can still exceed the offset capacity of the Gaussian primitives. We discuss these limitations and failure cases in the supplementary material.

\section{Conclusion and Limitations}

We presented DiGS-Avatar, an efficient diffusion framework that reformulates single-image 3D human reconstruction as 2D UV-latent completion. By combining a multi-view teacher-student distillation process with our semantic GASA module, the model generates photorealistic, instantly animatable 3D Gaussian avatars from a single uncalibrated image. Extensive evaluations demonstrate that DiGS-Avatar achieves state-of-the-art fidelity and robust zero-shot generalization while maintaining exceptional computational efficiency.

Despite its high efficiency and visual quality, DiGS-Avatar is fundamentally constrained by its reliance on the explicit SMPL-X topology. Because our geometric space is strictly bound to this template, the reconstruction quality can degrade when handling extreme, unconstrained poses or highly loose, non-rigid clothing that deviate significantly from the underlying body prior. Future work will focus on relaxing these strict topological constraints and exploring higher-resolution latents to better capture complex garments and fine-grained geometric topologies.

\section*{Acknowledgements}
This research was supported by the National Natural Science Foundation of China (NSFC) under Grant Nos. 62001432 and U25B2046, and the Fundamental Research Funds for the Central Universities under Grant No.~481014380011.

%
%
\bibliographystyle{splncs04}
\bibliography{main}
\clearpage
\input{supplementary}
\end{document}

%% file: supplementary.tex
\appendix

\setcounter{figure}{0}
\renewcommand{\thefigure}{S\arabic{figure}}

\setcounter{table}{0}
\renewcommand{\thetable}{S\arabic{table}}

\begin{center}
{\Large\bfseries DiGS-Avatar: Single-Image Animatable 3D Human Reconstruction via UV-Space Diffusion}\\[0.5em]
{\large \textit{Supplementary Material}}
\end{center}

\appendix

\section{Architectural Details}

\subsubsection{VAE Encoder.} 
We adopt the pretrained AE-F32C32P1 deep compression autoencoder (DCAE) from SANA \cite{xie2025sana} as our VAE encoder. All input images are first resized to $1024 \times 1024$ using aspect-preserving padding to avoid geometric distortion. The AE-F32C32P1 applies a 32× spatial downsampling and outputs 32 latent channels, converting each input image into a compact $32 \times 32 \times 32$ latent tensor. This latent tensor defines the unified representation space shared by both the teacher and student networks, and serves as the input domain for the diffusion model.

\subsubsection{Appearance Encoder.}
To capture high-frequency texture and semantically rich appearance cues, we adopt the pretrained DINOv3 ViT-B/16 encoder \cite{simeoni2025dinov3} as our appearance feature extractor. The encoder operates on $1024 \times 1024$ input images and processes them using a Vision Transformer with a $16 \times 16$ patch size, producing a dense feature map of resolution $64 \times 64$ with 768-dimensional tokens. This representation preserves mid- and high-level visual semantics---such as clothing patterns, shading, and fine structural cues---that are largely lost in the low-dimensional VAE latent. Since DINOv3 is trained with self-distillation and masked image modeling objectives, its features exhibit strong invariance to lighting and viewpoint while retaining rich texture information. These properties make it a powerful complement to the VAE latent, which provides geometric alignment but limited appearance detail. In our pipeline, the DINOv3 feature map is fused with the VAE latent through the GASA module to augment the UV representation with fine-grained appearance information in a geometry-aware manner.

\subsubsection{UV-Alignment Transformer.}
\begin{figure}[tb]
  \centering
  \includegraphics[width=\linewidth]{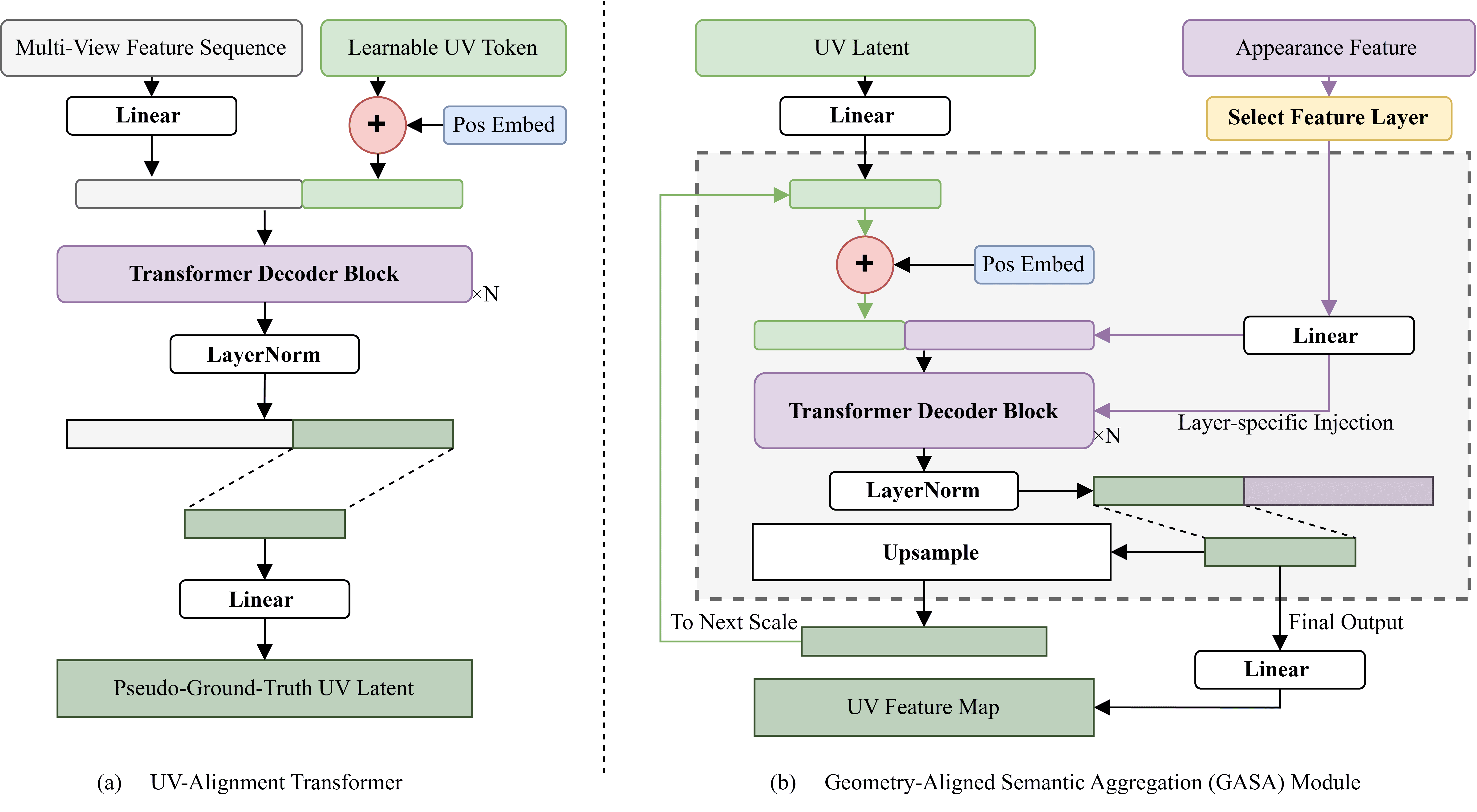}
  \caption{Detailed architecture of key modules: (a) UV-Alignment Transformer. (b) GASA module.
  }
  \label{fig:module_arch}
\end{figure}

Our UV-Alignment Transformer adopts a decoder-only architecture (see \cref{fig:module_arch}(a)) designed to aggregate features from multiple input images into a canonical $32 \times 32$ UV feature map. We first augment the flattened per-view image features with 6-dimensional Plücker ray embeddings---which provide explicit 3D spatial and camera perspective priors---and linearly project them to a unified hidden dimension. These context tokens are then concatenated with a set of 1,024 learnable query tokens, which are initialized with 2D sin-cos positional encodings to denote their structural coordinates in the target UV space. The combined token sequence is processed by a stack of 4 transformer blocks with an embedding width of 256. Each block follows a standard Pre-LN design, consisting of multi-head self-attention over the full sequence followed by a feed-forward network, with residual connections around both components. This global self-attention mechanism enables the UV query tokens to flexibly attend to all multi-view observations simultaneously, progressively integrating geometrically consistent features across different views. After the final layer, only the updated 1,024 query tokens are extracted, linearly projected to 32 channels, and reshaped to form the final $32 \times 32 \times 32$ UV feature map.

\subsubsection{Latent Diffusion Model.}
For diffusion-based UV-latent completion, we employ the SANA-1.6B latent diffusion transformer \cite{xie2025sana}--a 1.6B-parameter Linear DiT backbone. The model consists of 20 transformer blocks with width 2240, where all attention modules are replaced with efficient linear attention. A Mix-FFN module aggregates local information, enabling the model to operate without positional embeddings (NoPE). The diffusion model processes the VAE latent grid at its native spatial resolution: each element of the $32 \times 32$ latent map is treated as a $1 \times 1$ patch, resulting in a sequence of 1024 tokens, consistent with the original SANA tokenization. Text conditioning in SANA is unused in our setting; instead, the model is conditioned solely on geometric and appearance features extracted from the single input image. At inference, we use a Soft Blending Euler sampler with 20 denoising steps and CFG set to 0.1. During student fine-tuning, all non-diffusion modules initialized from the teacher are frozen, and only the diffusion model is updated. The loss weights are set to $\lambda_\text{vgg}=1.0$, $\lambda_\text{off}=0.5$, and $\lambda_\text{photo}=0.5$. The reported 0.71s runtime includes the full pipeline from input image to final 3DGS output.

\subsubsection{GASA module.}
In our framework, the Geometry-Aligned Semantic Aggregation (GASA) module bridges the 3D-consistent structural skeleton with high-level semantic cues. The term ``Geometry-Aligned'' refers to the intrinsic, 3D-aware structure of the latent space: during the teacher network's training, multi-view features are fused using explicit Plücker ray embeddings to resolve spatial consistency. While the student network does not use Plücker embeddings during single-view inference, it successfully distills this geometrically structured latent, ensuring that 2D appearance features are subsequently injected into a strictly consistent 3D skeleton without introducing geometric tearing. To enrich this skeleton, GASA utilizes appearance features extracted from 12 transformer layers of a pre-trained DINOv3 encoder, strategically divided into deep, middle, and shallow groups of four layers each. As illustrated in \cref{fig:module_arch}(b), the module employs a three-stage progressive upsampling mechanism to integrate these layer groups. It begins at a base spatial resolution of $32 \times 32$, where the input UV latent is augmented with explicit positional embeddings and processed through Transformer Decoder blocks to query the deep appearance features via cross-attention. Following a LayerNorm and upsample operation, the spatially expanded $64 \times 64$ latent enters the second stage to query the middle four DINOv3 layers, injecting mid-level structural and textural semantics. Finally, the latent is upsampled to a $128 \times 128$ resolution to query the shallow four layers, recovering the finest high-frequency textural details (e.g., fabric grain and complex clothing patterns) before a final linear layer produces the UV feature map for the 3DGS decoder.

\subsubsection{UV-to-3DGS Decoder.}
The UV-to-3DGS decoder converts the enriched UV feature tokens into dense Gaussian attribute maps through a hierarchical convolutional architecture. The UV tokens are first reshaped into a $512 \times 128 \times 128$ feature map, which serves as the canonical UV representation. This map is processed by a sequence of upsampling stages, each implemented as an interpolate-convolution block: bilinear interpolation followed by a $3 \times 3$ convolution, instance normalization, and a SiLU activation. These blocks progressively increase spatial resolution while refining local structure. After sufficient stages we obtain a $128 \times 1024 \times 1024$ UV feature map. A three-layer convolutional refinement module further enhances feature quality and produces the shared feature map for two separate convolutional heads. Following \cite{zhang20243gen}, we then employ two separate convolutional heads for geometry and appearance decoding. Each head progressively transforms the shared 32-channel feature map into task-specific outputs: the geometry head predicts the Gaussian parameters $(\delta \mu_k, \delta s_k, \delta r_k)$, while the color head predicts the appearance parameter $c_k$. This decoupled design improves network specialization and stability, enabling the decoder to produce accurate, high-fidelity canonical 3D Gaussian primitives suitable for differentiable rendering and animation.

\section{Detailed Baseline Configurations and Fairness Justification}
\label{supp:baselines}

To ensure a rigorously fair evaluation, all baseline methods evaluated in our study (IDOL, LHM, SIFU, Human3Diffusion, TRELLIS, and SyncHuman) were executed using their officially released pre-trained weights and recommended inference pipelines. For all dataset evaluations (HuGe100K, THuman 2.1, 2K2K, SIZER), every method was provided with the exact same ground-truth camera parameters and ground-truth SMPL-X fits. 

Our training protocol is largely aligned with IDOL, using HuGe100K and THuman2.1 with consistent splits. We additionally use 2K2K for final fine-tuning because our multi-view teacher requires geometrically consistent multi-view supervision; diffusion-generated views in HuGe100K may contain cross-view inconsistencies that harm teacher latent construction. In contrast, single-image regression methods such as IDOL do not rely on multi-view fusion during training. Since retraining all large baselines under identical data and compute budgets is impractical, we evaluate official released checkpoints. Thus, our comparison reflects the maximum achievable performance of each method under practical, real-world computational constraints.

\section{Additional Qualitative Results}

\subsection{Reconstruction on In-the-Wild Images}
\begin{figure}[tb]
    \centering
    \includegraphics[width=\textwidth]{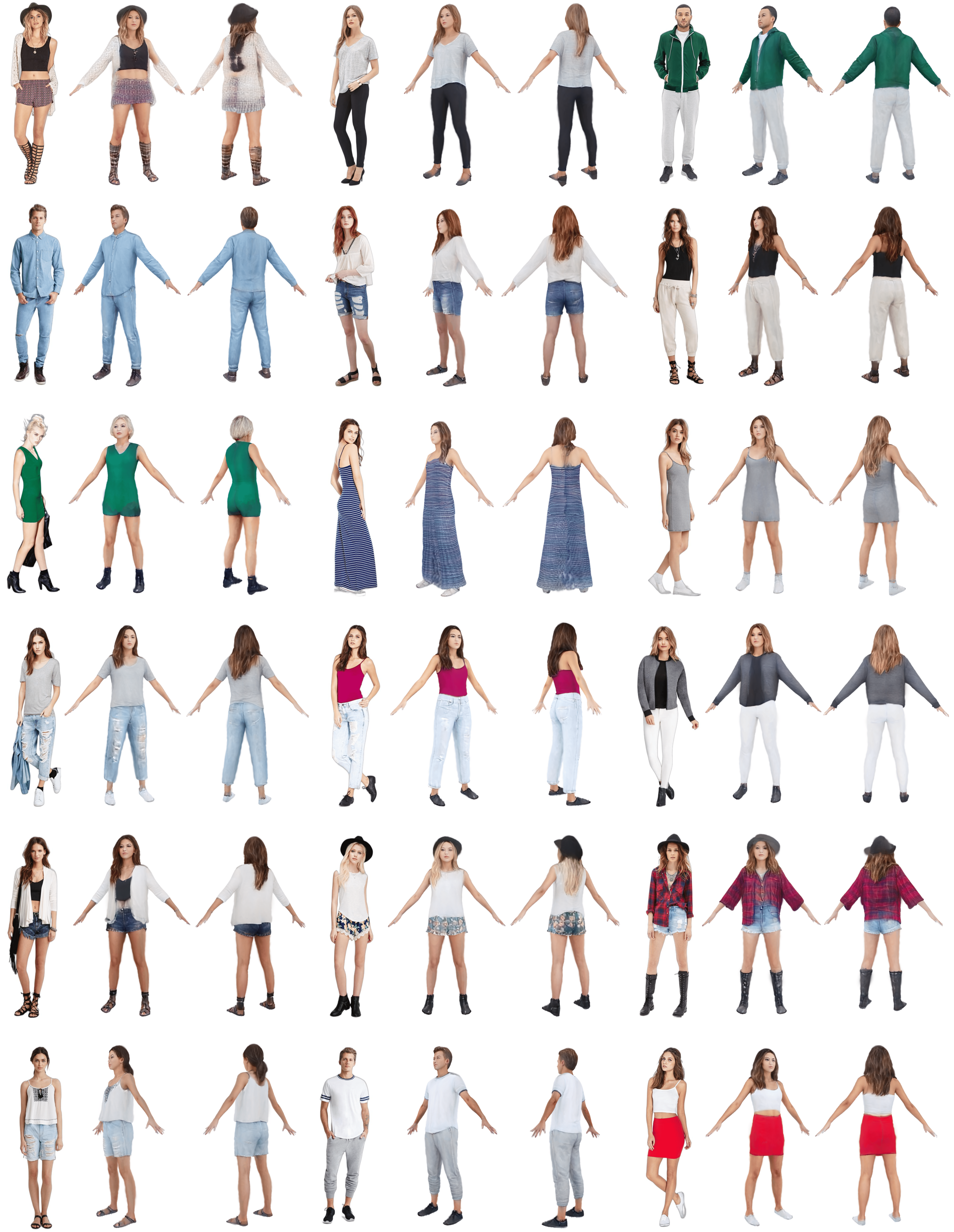}
    \caption{Additional in-the-wild reconstruction results under diverse clothing, poses, and illumination.}
    \label{fig:supp_inthewild}
\end{figure}

To further demonstrate the robust generalization capabilities of DiGS-Avatar, we provide extensive additional reconstruction results on unconstrained, in-the-wild images in \cref{fig:supp_inthewild}. These everyday photographs feature complex real-world illumination, heavy occlusions, and highly diverse casual garments (e.g., flowing coats, multi-layered dresses, and loose-fitting sportswear). By leveraging Multi-HMR for automatic SMPL-X parameter estimation, our framework successfully handles these uncalibrated inputs. The qualitative results clearly show that our geometry-aligned diffusion prior effectively hallucinates plausible textures for unseen back-views and preserves high-frequency structural details, demonstrating practical applicability far beyond the constrained studio environments of standard 3D scan corpora.

\subsection{Animation and Temporal Stability}
\begin{figure}[tb]
    \centering
    \includegraphics[width=\textwidth]{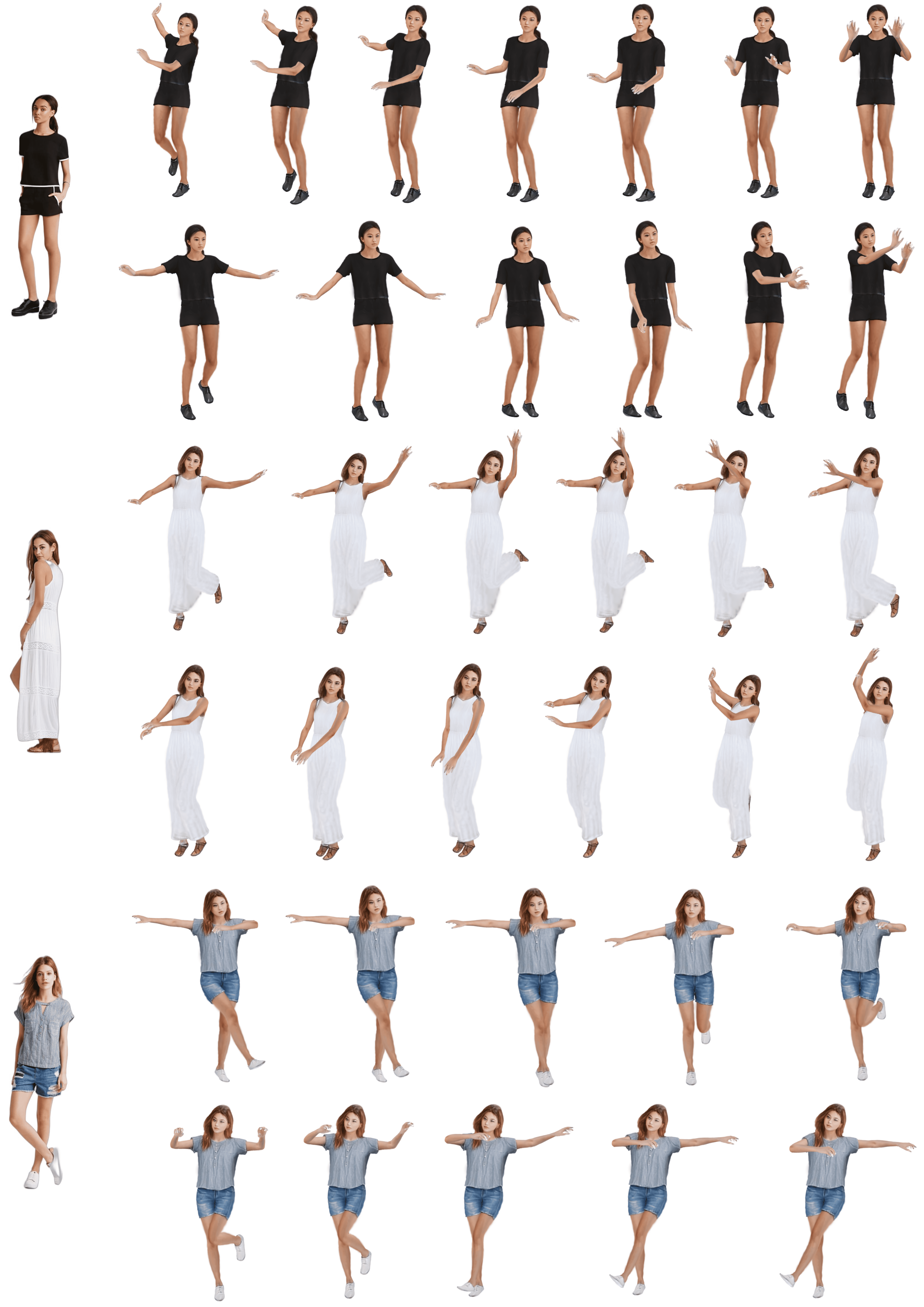}
    \caption{Additional animation results driven by novel pose sequences.}
    \label{fig:supp_animation}
\end{figure}

In \cref{fig:supp_animation}, we present additional frame-by-frame visualizations of our reconstructed avatars driven by novel pose sequences. Because DiGS-Avatar explicitly anchors the predicted 3D Gaussian primitives to the canonical SMPL-X UV space, the generated avatars are instantly animatable using standard Linear Blend Skinning (LBS), entirely bypassing the need for computationally expensive test-time optimization or the learning of complex deformation fields.

\clearpage

\section{User Study}
To rigorously evaluate the perceptual quality and practical animation stability of our method against state-of-the-art animatable baselines, we conducted a comprehensive user study following the evaluation protocols established in IDOL. We randomly selected 50 challenging single-view cases (spanning both dataset and in-the-wild domains). For each case, we generated 3D reconstructions and animation sequences using DiGS-Avatar and the two strongest animatable deterministic regressors: IDOL and LHM. We restrict the user study to directly LBS-animatable avatar baselines. Diffusion-native methods such as Human3Diffusion, TRELLIS, and SyncHuman are included in reconstruction comparisons, but require additional rigging or retargeting to follow the same animation protocol.

We recruited $N=30$ human evaluators, who were presented with the reference input image alongside rendered novel views and animated sequences from the three methods side-by-side in randomized order. Participants were asked to select their single preferred result across three distinct criteria, as well as an overall preference:
\begin{enumerate}
    \item \textbf{Geometric Consistency:} The structural accuracy of the reconstructed 3D shape and the plausibility of hallucinated unseen regions (e.g., the back of the subject).
    \item \textbf{Texture Clarity:} The sharpness of the generated textures, preservation of high-frequency details (e.g., facial features, clothing patterns), and overall photorealism.
    \item \textbf{Animation Quality:} The temporal coherence, structural stability, and absence of artifacts (e.g., texture sliding or geometric tearing) when the avatar is driven by novel poses.
\end{enumerate}

\begin{table}[h]
    \centering
    \caption{\textbf{User Preference Rates (Ours vs. IDOL \& LHM).} Percentage of times evaluators preferred DiGS-Avatar over the animatable baselines across 50 challenging cases ($N=30$).}
    \label{tab:user_study}
    \begin{tabular}{lcccc}
    \toprule
    Metric & Geometric Consistency & Texture Clarity & Animation Quality & \textbf{Overall} \\
    \midrule
    Preference & 66.6\% & 60.0\% & 56.6\% & \textbf{63.3\%} \\
    \bottomrule
    \end{tabular}
\end{table}

\section{Sensitivity to SMPL-X Estimation and Failure Cases}
\label{supp:limitations}

While DiGS-Avatar demonstrates strong zero-shot generalization on in-the-wild images by leveraging Multi-HMR for automated SMPL-X parameter estimation, its performance is inherently bounded by the accuracy of this underlying topological prior. When the estimated SMPL-X pose or shape is inaccurate, our method may avoid severe tearing but still suffers from feature misalignment and identity degradation, resulting in the following specific failure modes:

\begin{itemize}
    \item \textbf{Feature Misalignment and Blurring:} When the estimated SMPL-X pose or shape is inaccurate, the geometry-aligned UV latent is forced to sample from incorrect spatial locations in the reference image. Because the structural skeleton does not align with the visual evidence, the GASA module cannot accurately extract high-frequency semantic features. Consequently, specific attributes fail to match the input image (e.g., incorrect clothing colors or mismatched pants lengths), and critical identity-preserving details like the face and hairline become highly blurred.
    \item \textbf{Regression to the Mean Body Shape:} In cases of severe pose failure, the diffusion model lacks reliable geometric conditioning and thus falls back heavily on its learned generative prior. Instead of producing broken geometry, the network hallucinates a plausible but incorrect ``average'' human. This results in the reconstruction trending toward an average face and generic body shape, leading to visible discrepancies in the subject's body shape and volume, limb angles, and limb proportions compared to the actual reference image.
    \item \textbf{Extreme Non-Rigid Topologies:} Our method mitigates the strictness of the SMPL-X prior by predicting explicit 3D positional offsets ($\delta\mu_k$) for the Gaussian primitives, allowing them to deviate from the bare-body surface to capture standard clothing. However, in extreme cases---such as subjects wearing exceptionally loose, floor-length dresses---this offset capacity is exceeded. Because the primitives remain fundamentally anchored to the canonical UV topology, these massive deviations cannot be properly parameterized, often resulting in flattened textures or clipped geometry for the extreme non-rigid elements.
\end{itemize}

Relaxing these strict topological constraints to better capture unconstrained geometries and improving the robustness of the structural latent against initial pose errors remain primary directions for our future work.